%% file: paper.tex
\documentclass[sigconf, nonacm]{acmart}
\usepackage{fancyhdr}
\renewcommand\footnotetextcopyrightpermission[1]{} 
\usepackage{amsmath}

\usepackage{multirow}
\usepackage{array}
\usepackage{epsfig}
\usepackage{subfig}
\usepackage{graphicx}  
\usepackage{float}  
\usepackage{multirow}
\usepackage{array}
\AtBeginDocument{%
  \providecommand\BibTeX{{%
    \normalfont B\kern-0.5em{\scshape i\kern-0.25em b}\kern-0.8em\TeX}}}

\setcopyright{acmcopyright}
\copyrightyear{2020}
\acmYear{2020}
\acmDOI{10.xxxx/xxxxxxx.xxxxxxx}

\begin{document}

\title{Predicting Disease Progress with Imprecise Lab Test Results
}

\author{Mei Wang}
\authornote{A part of work is done while visiting UCSB.}
\author{Zhihua Lin}
\affiliation{%
  \institution{Donghua University, China}
}
\email{wangmei@dhu.edu.cn}

\author{Jianwen Su}
\affiliation{
  \institution{Univ of California, Santa Barbara, USA}
}
\email{su@cs.ucsb.edu}

\renewcommand{\shortauthors}{}

\begin{abstract}
In existing deep learning methods,
almost all loss functions assume
that sample data values used to be predicted 
are the only correct ones. 
This assumption does not hold
for laboratory test data. 
Test results 
are often within tolerable or imprecision ranges,
with all values in the ranges acceptable. 
By considering imprecision samples,
we propose an imprecision range loss (IR loss) method
and incorporate it into Long Short Term Memory (LSTM) model for
disease progress prediction.
In this method, 
each sample in imprecision range space
has a certain probability 
to be the real value, 
participating in the loss calculation.
The loss is defined 
as the integral of the error 
of each point 
in the impression range space.
A sampling method 
for imprecision space 
is formulated. 
The continuous imprecision space 
is discretized, and 
a sequence of imprecise data sets are obtained,
which is convenient 
for gradient descent learning.
A heuristic learning algorithm 
is developed to 
learn the model parameters 
based on the imprecise data sets.
Experimental results 
on real data show 
that the prediction method based on 
IR loss can provide more stable and consistent prediction result 
when test samples are generated from imprecision range.
\end{abstract}



\maketitle

\input sec1-intro

\input sec2-related

\input sec3-method

\input sec4-expeval

\input sec5-con

\begin{acks}
This work 
by 
Wang and Lin was supported 
by 
the National Key R\&D Program 
of
China under Grant 2019YFE0190500.
\end{acks}

\input{MaggieRef.bbl}

\end{document}

%% file: sec1-intro.tex
\section{Introduction}
In healthcare, it is highly desirable to 
evaluate the current situation of 
a patient and 
predict his/her disease development. 
This evaluation and prediction
provide a basis for treatments 
including the medication strategy, non-routine checkup, 
early active interventions, etc.
Deep neural networks (DNNs)
have been increasingly applied to
the prediction, prevention, diagnosis and 
prognosis of diseases, leading to
a promising capability
for better decision-making \cite{nature}\cite{Medical}.
Although DNNs have proven their merits 
in various tasks,
the performance in noise, disturbance and imprecise data
remains a challenge. 
Obtaining more stable and robust medical deep learning models
is at the forefront of machine learning in healthcare.

Clinical lab tests play an important role
in healthcare.
From early detection of diseases
to diagnosis to personalized treatment programs,
lab tests guide more than 70\% of medical decisions
and personalized medication \cite{labtesting14}.
However,
due to limitations of equipment, instruments, materials,
test methods, etc.,
data inaccuracy always occurs \cite{CV14}\cite{allowableimp} and has
to be dealt with. 
Typically, test results 
are within respective tolerable ranges 
(or imprecision ranges, 
values in this range are acceptable 
though imprecise). 
In our previous work \cite{ourwork}, we studied
the impact of imprecision on
prediction results 
where a pre-trained model is used
to predict future state of hyperthyroidism for patients.
It was demonstrated that small ranges of imprecisions
can cause large ranges of predicted results,
which might cause mis-labeling and
inappropriate actions (treatments or no treatments)
for individual patients. 
In this paper, 
we study the issue of
building robust models
while taking imprecision into account 
with better generalization.

In image classification, 
it has long been found that
if one image is visually imperceptibly perturbed, 
a prediction label may be different.
To overcome this problem,
a popular solution is 
noise injection \cite{noiseinjection}.
The main focus was on 
methods to generate useful noisy 
or adversarial examples.
By including noise-injected examples in training,
resulting classification models are more sensitive to
discriminate these images.
In healthcare, 
adversarial patients have also been studied 
in different kinds of classification task \cite{Qayyum}\cite{adverclinical}.  
Obviously, samples in imprecision range 
are not noise, 
since imprecise samples 
are also accepted values.
The above approaches need to be extended 
to address
imprecision range problem readily. 

In existing deep learning methods,
loss functions play a central role 
in model learning. 
By calculating the local gradient 
of a loss function,
the gradient descent (GD) algorithm 
updates the model parameters 
in each iteration.
Virtually all loss functions assume that the values
in a learning dataset
are the only correct values.
A predicted value based on these is then used
to calculate the loss
for gradient descent. 
However, in lab tests for patients,
the values in imprecision range 
usually have some probability of true values,
By assuming the lab test results have the only correct values,
the learned model will likely deviate from the real model,
leading to incorrect, inconsistent predictions
when predicting new samples. 

In this paper, 
``IR loss'' is proposed 
and incorporated into LSTM model for
disease progress prediction.
In our method, 
each data in imprecision range space 
has a certain probability 
to be the real value, 
participating in the loss calculation.
So the loss is defined 
as the integral of the error 
of each point 
in the impression range space.
Then the sampling method 
for imprecision space 
is designed. 
The continuous imprecision space 
is discretized, 
and 
a sequence of imprecise data sets 
are obtained.
Then a heuristic learning algorithm 
is proposed to 
learn the model parameters 
based on the imprecise data sets 
sequentially.
Experimental results 
on a real dataset show 
that the prediction method based on IR loss is more robust,
which can provide more stable and consistent prediction result
when test samples are generated from imprecision range.

The 
paper is organized as follows.
Section 2
discusses related work.
Section 3
introduces the proposed loss function 
and the corresponding model learning method.
Section 4
presents experimental results.
Section 5 concludes the paper.

%% file: sec2-related.tex
\section{Related Work}

One area of work 
studied the influence of
variety forms of input perturbations
through experimental or theoretical analysis.
In \cite{SensitivityNPL}, 
an iterative algorithm  
was developed for approximately 
compute the sensitivity 
from neuron to layer, 
and finally to the entire CNN network.
Impact of data precision on learning
was first
observed in our earlier paper \cite{ourwork},
which also discussed related work.
This paper continues the study 
by developing an effective learning method.

To learning against input perturbations, 
a popular method is
noise injection \cite{CoRR}\cite{noiseinjection}.
By adding the noisy or adversarial 
samples into training process,
the model is expected to have more distinguish ability 
in classification task.
The studies focused more on to generating such samples.
In healthcare field,
adversarial patient 
are defined and examined in \cite{adverclinical}.
Different methods \cite{poisoningattack}\cite{advmedical} were provided 
to generate adversarial examples
in medical deep learning classifiers.
Obviously, we cannot 
treat each samples in imprecision range as noise, 
since imprecise samples
could be accepted values.
So the current solution 
cannot be used into
imprecision problem readily. 
How to use unseen imprecise samples 
to build a more robust model 
is addressed in this paper.

For discretize the continuous input space,
interval valued data regression method 
provides a potential solution.
In \cite{intervallearning},
each interval valued observation was 
viewed as a hypercube. 
By dividing each side of a $p$-dimensional
hypercube into $m$ equal parts 
to discretize the interval data.
However, 
the data in this paper 
is essentially different 
from interval data.
In the interval method, 
each data value is an interval
(e.g., cluster data) represented as 
a minimal and a maximal value.
In this paper, 
a test result is single-valued element, 
but its value can be possibly anyone 
in imprecision range.

%% file: sec3-method.tex
\section{Method}

In this paper, 
a prediction model $f$ 
is developed to 
predict the progress of hyperthyroidism 
two years in advance 
based on the blood test data 
in the first $k$ months. 
That is $y \,{=}\, f(x)$, 
where $x$ is the blood test result sequence
in the first $k$ months. 
$y$ is the predicted values two years later. 
After we obtain the predicted test values, 
the disease progress is obtained
by analyzing the ``normal'', ``abnormal'' states 
of the test measures.
Since $x$ can be regarded as a time series, 
we exploit LSTM 
as our basic prediction model. 
In the traditional LSTM modeling 
and training process, 
SGD method are often used to 
learn the model parameters
based on an loss function.
The traditional loss 
for a training sample $x$ 
is:
\begin{equation}
    L(x) = |y - f(x)|^\alpha
    \label{eqn_prevloss}
\end{equation}
where $\alpha$ is 1 or 2,
corresponding to the commonly used
absolute loss function 
and square loss function.

In Equation\,(\ref{eqn_prevloss}),
$x$ is the group of 
blood test results. 
Each test value
has an imprecision range,  
any value from this range 
is also acceptable.
Obviously, 
the loss defined in Equation\,(\ref{eqn_prevloss})
does not consider the samples 
in the imprecision range. 
It simply assumes 
the lab test value
is the only correct one. 
The predicted result 
based on this value is used to 
calculate the loss 
and gradient. 
If it is not the correct value,
the above calculation may 
cause the deviation in parameter learning
and correspondingly the inaccurate, inconsistent 
prediction results for the test data. 
\begin{figure}
\centering
\includegraphics[width=5cm]{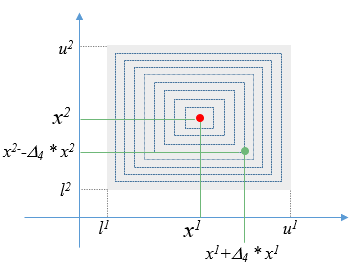}
\vspace*{-3mm}
\caption{The discretization example in two dimensional space. }
\label{fig:discre}
\vspace*{-3mm}
\end{figure}

\def\Ipr{\textit{Ipr}}

To deal with the above problem, 
we modify the imprecision range loss. 
Let $\Ipr(x)$ denote the imprecision range for $x$.
Assume data $x'\,{\in}\, \Ipr(x)$ is the true value 
with a certain probability $p(x')$, 
then the loss at $x$ could be defined as:
\begin{equation}
  L_D(x) = \int_{\Ipr(x)}\! P(x) L(x)dx
  \label{eqn:LD}
\end{equation}
where
$P(x)$ is the probability 
of $x$ being the true value such that
$\int_{\Ipr(x)}\!P(x)dx = 1$.
The 
probability $P(x)$
can also be regarded as the weight 
to control each samples 
in imprecision range 
contributed to the total loss of $x$.
When $P(x)\,{=}\, 1$ and $P(x')\,{=}\, 0$
for each $x'$ in $\Ipr(x)$ and $x' \neq x$,
Equation\,(\ref{eqn:LD}) degenerates to the traditional loss function
$L(x)$
in Equation\,(\ref{eqn_prevloss}).
In this case, $x$ is the only correct value.

The total loss for all training samples is given as:
\begin{equation}
   Loss = \sum  L_D(x) = \sum_{x \in D} \int_{\Ipr(x)}\! P(x) L(x)dx,
\end{equation}

The right side of the above equation
is a continuous integral term, and is possibly hard to evaluate.
We consider to simplify it by discretization.
Since each $x$ is a multivariate variable. 
Its imprecision ranges form
an imprecision space with infinitely dense points. 
We use the points with fixed size steps to discrietize this space.

Let $x^k$ denote a single blood test result,
$r$ denote its maximum acceptable range 
as a percentage. 
The lower and the upper boundary 
are then $x^k \pm r \,{*}\, x^k$. 
We uniformly generate the imprecise sample $x^k$
with fixed step $s$. 
The sample scale 
is defined as $\Delta_j = \Delta_0 + (j{-}1)*s$, 
where $\Delta_0 \,{=}\, 0$, $j \,{=}\, \{1, 2, ..., N\}$,
$N$ is the total number of steps. 
Obviously, we have $N = r/s$. 
For example,
if $r = 10\%, s=0.01$, 
we have $N\,{=}\,10$,
and $\Delta = \{0.01, 0.02, ..., 0.1\}$.  
This process can be easily 
extended to multidimensional input. 
Fig.\,1 illustrates an example 
in a two-dimensional space. 
$l^1, u^1, l^2, u^2$ are the lower and
upper boundaries of the imprecision space for
the two dimensional input value $x=\{x^1, x^2\}$.
The green dot is the generated imprecision sample when $\Delta = 0.04$.
In this way, 
the discretized imprecision sample $x_j$ 
is generated from its imprecision range 
as follows:
\begin{equation}
  x_j = x \pm \Delta_j  * x
\label{eqn:x'}
\end{equation}
For each sample $x$,
we generate 
a set of imprecision samples $\{x_1, x_2,$ $..., x_N\}$  
from its imprecision space. 
Let $x^0$  
be the lab returned value $x$. 
We further assume samples 
by using the same scale $\Delta$ share the same weight.
We discretize the weight vectors $\{w_0, w_1, ..., w_N\}$ 
from function $P(x)$.
Then the loss could be updated as:
\begin{eqnarray}
  Loss = \sum_{x \in D}  L_D(x) & = & \sum_{x \in D}\,  \sum_{x_i \in\Ipr(x)} P(x_i) L(x_i) \\
                               & = &\sum_{x \in D} \sum_{j=0}^{N} w_j L(x_j)
  \label{eqn:sum}
\end{eqnarray}
Equation\,(\ref{eqn:sum}) is also rewritten 
as $\sum_{x \in D} w_0 L(x) +\sum_{x \in D}\! \sum_{j=1}^{N}\! w_j L(x_j)$. 
The term $\sum_{j=1}^{N} w_j L(x_j)$ carries two meanings. 
The prediction based on each accepted samples
in the imprecision range 
should be as centralized
and close as possible. 
At the same time, 
all predictions should be 
as close as possible to the true value. 
It implies that the learning results 
should be accurate and stable, 
and consequently more trustable.

Further,
the imprecision samples 
with the same sample scale $\Delta_j$ 
are grouped together to form the data set $D_j$. 
Then the loss could be rewritten as:
\begin{eqnarray}
   Loss\!\! & = & \!\!\!\!\sum_{x_0 \in D_0} w_0 L(x_0)+\!  \sum_{x_1 \in D_1} w_1 L(x_1) +\! \ldots + \!\sum_{x_N \in D_N} w_N L(x_N),\nonumber \\  
 \!\!&  =  & \!\!\!\! \sum_{j = 0}^{N} Loss_{D_j}
   \label{eqn:sumLD}
\end{eqnarray}
where $Loss_{D_j} =  \sum_{x_j \in D_j} w_j L(x_j)$.

The goal of model training is to find the optimal model parameters
$\hat{\theta}$ to minimize the loss function defined in Equation\,(\ref{eqn:sumLD}):
\begin{eqnarray}
  \hat{\theta} & = & arg min_\theta\,\,\, Loss
   \label{eqn:theta}
\end{eqnarray}

To achieve this goal, 
we divide the global
optimization problem 
into the combination 
of optimization problems 
on each individual imprecision set $D_i$.
It is reasonable to assume 
that the closer 
to the center point $x$,
the greater the weight should be. 
So we have $w_0 \textgreater w_1 \textgreater w_2 \textgreater \cdots \textgreater w_N$. 
The model is trained sequentially 
from larger weights to smaller ones. 
Specifically, 
the model is trained on the first data set $D_0$ 
to obtain the initial parameters 
by minimizing $Loss_{D_j}$. 
Then, on the basis of the initial model, 
we use $D_1$ for the next training round to get model $f_1$.
The above process is carried out in turn. 
The final prediction model $f_N$ is obtained 
after $N$ iteration. 
The detailed learning process is illustrated 
in Algorithm 1.

\begin{table}[htbp]
\centering
\begin{tabular}{p{0.5cm}p{7cm}}
\hline
\multicolumn{2}{l}{\textbf{Algorithm 1: Model learning}}\\
\hline
\multicolumn{2}{l}{\textbf{Input:} training set }\\
\multicolumn{2}{l}{\textbf{Output:} model $\theta$}\\
0.& SGD learning $\hat{\theta}_0$ by minimizing the loss $Loss_{D_0}$; \\
1.& for each  $D_i \in \{D_1, D_2, ..., D_N\}$;\\
2.&  \hspace*{0.5cm} Initial model parameter using $\theta_{i-1}$;\\
3.&  \hspace*{0.5cm} SGD training new $\theta$ to obtain $f_i$;\\
4.&  \hspace*{0.5cm} update model $f = f_i$;\\
5.& end for \\
6.& return finial model ${f_N}$ \\
\hline
\end{tabular}
\vspace*{0mm}
\end{table}

%% file: sec4-expeval.tex
\begin{figure*}
\centering
\begin{minipage}[t]{0.48\textwidth}\centering
\includegraphics[width=7.8cm]{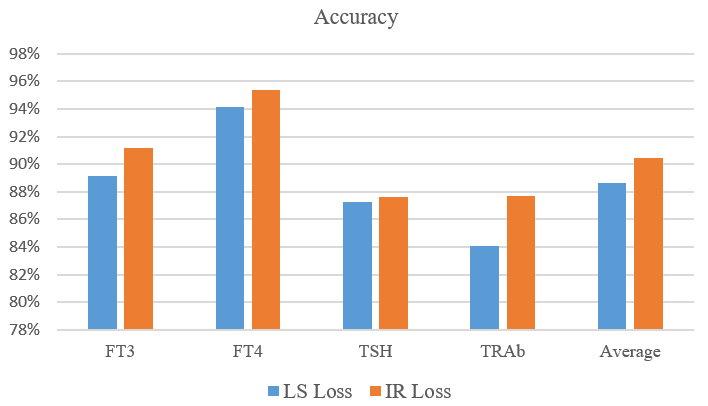}
\vspace*{0mm}  
\caption{The $Accuracy$ comparison of LS loss and IR Loss}
\label{fig:accuracy}
\vspace*{0mm}
\end{minipage}
\begin{minipage}[t]{0.48\textwidth}\centering
\includegraphics[width=7.8cm]{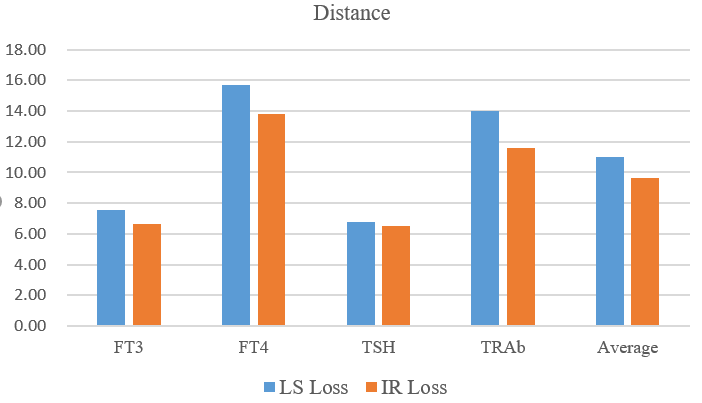}
\vspace*{0mm}  
\caption{The $Distance$ comparison of LS loss and IR Loss}
\label{fig:loss}
\end{minipage}
\vspace*{0mm}
\begin{minipage}[t]{0.48\textwidth}\centering
\includegraphics[width=7.8cm]{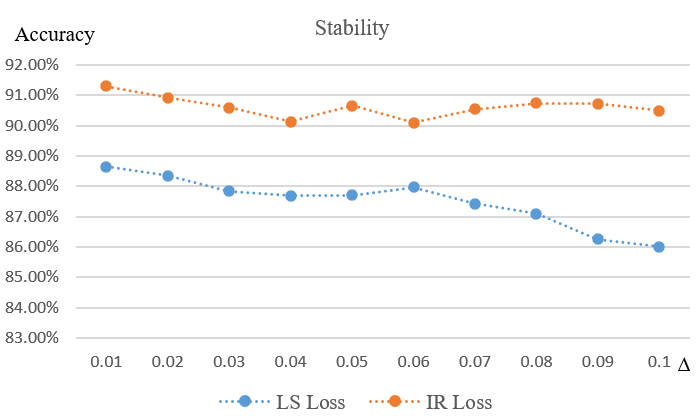}
\vspace*{0mm}  
\caption{The prediction stability of the proposed IR Loss}
\label{fig:stability}
\vspace*{0mm}
\end{minipage}
\begin{minipage}[t]{0.48\textwidth}\centering
\includegraphics[width=7.8cm]{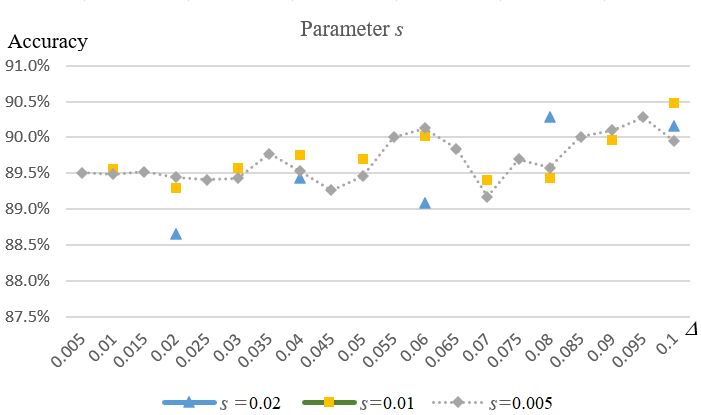}
\vspace*{0mm}  
\caption{The performance with different parameter setting }
\label{fig:para_s}
\end{minipage}
\vspace*{0mm}
\end{figure*}

\section{Experimental Evaluations}

\subsection{Experimental Setup}

There are 2,460 patients in the hyperthyroidism dataset used in our experiments, which
is is generated from Ruijin hospital, 
a reputable hospital in Shanghai, China.
The whole dataset is divided into a training set and a test set,
with 1960 patients and 500 patients respectively. For hyperthyroidism disease, the blood test measures FT3, FT4, TSH, and TRAb are used for prediction.

In the LSTM network, there are 2 hidden layers,
each layer containing 128 LSTM cell units.
We employ dropout method to reduce over-fitting and
apply the Adam-Optimizer in training.
Each experiment runs for 10 times and each data given
in the experimental results is the average of the 10 runs.
For the parameter setting, $r$ is set to be $10\%$ and $s=0.1$,
then $\Delta$ is 0.01 $\thicksim$ 0.1 with interval 0.01 to generate $D_1 \thicksim D_{10}$.

\subsection{Evaluation Metrics}

After predicting $f(x)$,
the label $l(f(x))$ for ``normal'', ``abnormal'' description can be obtained by
comparing the value of $f(x)$ with the reference range.
We use $Accuracy$ to measure the label prediction accuracy of the prediction method. Given one label $l$, let $\textit{TP}$
model the positive samples in the test set and
the predictions are correct.
$\textit{TN}$ represents the negative samples and the predictions
are correct. For each test patient $x$, we generate the imprecise sample $x_j, j=\{1, 2, \ldots, N\}$ from its imprecision range. In order to measure the value prediction accuracy, we further define the $Distance$ metric, which is the distance from the predicted value based on imprecision samples to the real value. 
$$
\textit{Accuracy}  =  \frac{|\textit{TP}\,| + |\textit{TN}\,|}{|D_{\textrm{Test}}|}
$$
$$
\textit{Distance} = \sum_{j=1}^N |f(x_j) - y|^{\alpha}
$$

\subsection{Findings}
In experiments, 
we compare the performance
of the new Imprecision Range Loss (IR Loss)
with the one 
by adopting the traditional Least Square Error loss (LS Loss).

Fig.\,\ref{fig:accuracy}  illustrates
the $Accuracy$ comparisons.
We can see that
the average $Accuracy$ 
and the $Accuracy$ of all the four test measures 
based on the proposed method 
are higher 
than the traditional ones. 
Especially for the TRAb measure, 
there are 4\% improvements. 

From the definition, 
the metric $Distance$ describes 
the degree of deviation 
between the predicted values and 
the true values 
when $x$ are 
sampled from the imprecision range.
According to Fig.\,\ref{fig:loss}, 
we can see that
the $Distance$ of the proposed method
are all smaller than the traditional method. 
It demonstrates that the method
based on the proposed IR loss could provide
more consistent prediction 
in the small range.   

Our previous work demonstrated 
that the performance declines 
when test samples
are generated 
from its imprecision range \cite{ourwork}.
To see how the proposed IR loss 
address this problem,
in Fig.\,\ref{fig:stability}, 
we provide the results 
when the proposed method 
applied to imprecise test data
when $Delta$ is set to 
be different values. 
From the figure, 
we can see that 
the accuracy of
the proposed IR loss is always higher 
than that of traditional loss. 
In addition, when $\Delta$ increases, 
the decrease rate of $Accuracy$ 
of the proposed method 
is much lower than 
that of the traditional method. 
This trend is more obvious
when $\Delta$ is large.  

Discretization is very important 
in the proposed method.
In this experiment,
 Discretization parameter $s$
is set to be 0.02, 0.01, 0.005 respectively.
Then three different $\Delta$ sequences obtained.
The prediction performance 
in different sampling sequences 
is illustrated in Fig.\,\ref{fig:para_s}.
The figure
does not establish that
finer the sampling granularity provides
better results. 
When $s=0.01$, 
the method achieve the best performance.
Similarly, 
it is not clear that 
the more imprecise data sets 
are introduced, 
the better the performance will be.
The increase of $\Delta$  means the number of imprecise data set increases, 
the performance does not steadily increase.

%% file: sec5-con.tex
\section{Conclusions}

The learning method presented in this paper
centers around a refined error function and
input data imprecision range.
This basically addressed the problems observed in our previous study
\cite{ourwork} with a satisfactory solution.
Further improvements
may possibly take into consideration of
output data imprecision range
and 
probability distributions
of the imprecise samples.
Also, it remains to be seen if such methods
are effective in other applications
in, e.g., healthcare and engineering.

%% file: MaggieRef.bbl

%% file: paper.bbl

\begin{thebibliography}{14}


\ifx \showCODEN    \undefined \def \showCODEN     #1{\unskip}     \fi
\ifx \showDOI      \undefined \def \showDOI       #1{#1}\fi
\ifx \showISBNx    \undefined \def \showISBNx     #1{\unskip}     \fi
\ifx \showISBNxiii \undefined \def \showISBNxiii  #1{\unskip}     \fi
\ifx \showISSN     \undefined \def \showISSN      #1{\unskip}     \fi
\ifx \showLCCN     \undefined \def \showLCCN      #1{\unskip}     \fi
\ifx \shownote     \undefined \def \shownote      #1{#1}          \fi
\ifx \showarticletitle \undefined \def \showarticletitle #1{#1}   \fi
\ifx \showURL      \undefined \def \showURL       {\relax}        \fi
\providecommand\bibfield[2]{#2}
\providecommand\bibinfo[2]{#2}
\providecommand\natexlab[1]{#1}
\providecommand\showeprint[2][]{arXiv:#2}

\bibitem[\protect\citeauthoryear{??}{CV1}{2014}]%
        {CV14}
 \bibinfo{year}{2014}\natexlab{}.
\newblock \bibinfo{title}{Desirable Biological Variation Database
  specifications}.
\newblock
\newblock
\urldef\tempurl%
\url{https://www.westgard.com/biodatabase1.htm}
\showURL{%
\tempurl}


\bibitem[\protect\citeauthoryear{??}{lab}{2014}]%
        {labtesting14}
 \bibinfo{year}{2014}\natexlab{}.
\newblock \bibinfo{title}{Importance of Clinical Lab Testing Highlighted During
  Medical Lab Professionals Week}.
\newblock
\newblock
\urldef\tempurl%
\url{https://www.acla.com/importance-of-clinical-lab-testing-highlighted-during-medical-lab-professionals-week/}
\showURL{%
\tempurl}


\bibitem[\protect\citeauthoryear{et~al}{et~al}{2018}]%
        {Medical}
\bibfield{author}{\bibinfo{person}{S.~M.~Mckinney et al}.}
  \bibinfo{year}{2018}\natexlab{}.
\newblock \showarticletitle{Artificial intelligence will soon change the
  landscape of medical physics research and practice}.
\newblock \bibinfo{journal}{\emph{Medical physics}} (\bibinfo{year}{2018}),
  \bibinfo{pages}{1791–1793}.
\newblock


\bibitem[\protect\citeauthoryear{et~al}{et~al}{2020}]%
        {nature}
\bibfield{author}{\bibinfo{person}{S.~M.~Mckinney et al}.}
  \bibinfo{year}{2020}\natexlab{}.
\newblock \showarticletitle{International evaluation of an AI system for breast
  cancer screening}.
\newblock \bibinfo{journal}{\emph{nature}} (\bibinfo{year}{2020}),
  \bibinfo{pages}{89–94}.
\newblock


\bibitem[\protect\citeauthoryear{Finlayson, Chung, Kohane, and Beam}{Finlayson
  et~al\mbox{.}}{2018}]%
        {advmedical}
\bibfield{author}{\bibinfo{person}{S.~G. Finlayson}, \bibinfo{person}{H.~W.
  Chung}, \bibinfo{person}{I.~S. Kohane}, {and} \bibinfo{person}{A.~L. Beam}.}
  \bibinfo{year}{2018}\natexlab{}.
\newblock \showarticletitle{Adversarial attacks against medical deep learning
  systems}. In \bibinfo{booktitle}{\emph{arXiv:1804.05296}}.
\newblock


\bibitem[\protect\citeauthoryear{He, Rakin†, and Fan}{He
  et~al\mbox{.}}{2019}]%
        {noiseinjection}
\bibfield{author}{\bibinfo{person}{Z.~Z. He}, \bibinfo{person}{A.~S. Rakin†},
  {and} \bibinfo{person}{D.~L. Fan}.} \bibinfo{year}{2019}\natexlab{}.
\newblock \showarticletitle{Parametric Noise Injection: Trainable Randomness to
  Improve Deep Neural Network Robustness against Adversarial Attack}. In
  \bibinfo{booktitle}{\emph{IEEE Conference on Computer Vision and Pattern
  Recognition}}. \bibinfo{pages}{588--597}.
\newblock


\bibitem[\protect\citeauthoryear{Mozaffari-Kermani, Sur-Kolay, Raghunathan, and
  Jha}{Mozaffari-Kermani et~al\mbox{.}}{201}]%
        {poisoningattack}
\bibfield{author}{\bibinfo{person}{M. Mozaffari-Kermani}, \bibinfo{person}{S.
  Sur-Kolay}, \bibinfo{person}{A. Raghunathan}, {and} \bibinfo{person}{N.~K.
  Jha}.} \bibinfo{year}{201}\natexlab{}.
\newblock \showarticletitle{Parametric Noise Injection: Trainable Randomness to
  Improve Deep Neural Network Robustness against Adversarial Attack}.
\newblock \bibinfo{journal}{\emph{IEEE journal of biomedical and health
  informatics}} (\bibinfo{year}{201}), \bibinfo{pages}{1893–1905}.
\newblock


\bibitem[\protect\citeauthoryear{Papangelou, Sechidis, Weatherall, and
  Brown}{Papangelou et~al\mbox{.}}{2018}]%
        {adverclinical}
\bibfield{author}{\bibinfo{person}{K. Papangelou}, \bibinfo{person}{K.
  Sechidis}, \bibinfo{person}{J. Weatherall}, {and} \bibinfo{person}{G.
  Brown}.} \bibinfo{year}{2018}\natexlab{}.
\newblock \showarticletitle{Toward an understanding of adversarial examples in
  clinical trials}.
\newblock \bibinfo{journal}{\emph{in Joint European Conference on Machine
  Learning and Knowledge Discovery in Databases}} (\bibinfo{year}{2018}),
  \bibinfo{pages}{35--51}.
\newblock


\bibitem[\protect\citeauthoryear{Qayyum, Qadir, Bilal, and Al-Fuqaha}{Qayyum
  et~al\mbox{.}}{2021}]%
        {Qayyum}
\bibfield{author}{\bibinfo{person}{A. Qayyum}, \bibinfo{person}{J. Qadir},
  \bibinfo{person}{M. Bilal}, {and} \bibinfo{person}{A. Al-Fuqaha}.}
  \bibinfo{year}{2021}\natexlab{}.
\newblock \showarticletitle{Secure and Robust Machine Learning for Healthcare:
  A Survey}. In \bibinfo{booktitle}{\emph{arXiv:2001.08103}}.
\newblock


\bibitem[\protect\citeauthoryear{Szegedy, Zaremba, Sutskever, Bruna, Erhan,
  Goodfellow, and Fergus}{Szegedy et~al\mbox{.}}{2013}]%
        {CoRR}
\bibfield{author}{\bibinfo{person}{C. Szegedy}, \bibinfo{person}{W. Zaremba},
  \bibinfo{person}{I. Sutskever}, \bibinfo{person}{J. Bruna},
  \bibinfo{person}{D. Erhan}, \bibinfo{person}{I.~J. Goodfellow}, {and}
  \bibinfo{person}{R. Fergus}.} \bibinfo{year}{2013}\natexlab{}.
\newblock \showarticletitle{Intriguing properties of neural networks}. In
  \bibinfo{booktitle}{\emph{CoRR}}.
\newblock


\bibitem[\protect\citeauthoryear{Tim, Mohamed, Alexandra, and Andy}{Tim
  et~al\mbox{.}}{2018}]%
        {intervallearning}
\bibfield{author}{\bibinfo{person}{P. Tim}, \bibinfo{person}{Z. Mohamed},
  \bibinfo{person}{B. Alexandra}, {and} \bibinfo{person}{N. Andy}.}
  \bibinfo{year}{2018}\natexlab{}.
\newblock \showarticletitle{High-quality prediction intervals for deep
  learning: a distribution-free, ensembled approach}. In
  \bibinfo{booktitle}{\emph{Proceedings of the 35th International Conference on
  Machine Learning}}.
\newblock


\bibitem[\protect\citeauthoryear{Wang, Su, and Lu}{Wang et~al\mbox{.}}{2020}]%
        {ourwork}
\bibfield{author}{\bibinfo{person}{M. Wang}, \bibinfo{person}{J.~W. Su}, {and}
  \bibinfo{person}{H.~Q. Lu}.} \bibinfo{year}{2020}\natexlab{}.
\newblock \showarticletitle{Impact of Medical Data Imprecision on Learning
  Results}. In \bibinfo{booktitle}{\emph{Proceedings of SIGKDD workshop}}.
\newblock


\bibitem[\protect\citeauthoryear{Xiang, Zeng, Wu, Liu, and Yuan}{Xiang
  et~al\mbox{.}}{2021}]%
        {SensitivityNPL}
\bibfield{author}{\bibinfo{person}{L. Xiang}, \bibinfo{person}{X.~Q. Zeng},
  \bibinfo{person}{S.~L. Wu}, \bibinfo{person}{Y.~J. Liu}, {and}
  \bibinfo{person}{B.~H. Yuan}.} \bibinfo{year}{2021}\natexlab{}.
\newblock \showarticletitle{Computation of CNN’s Sensitivity to Input
  Perturbation}.
\newblock \bibinfo{journal}{\emph{Neural Processing Letters}}
  \bibinfo{volume}{53} (\bibinfo{year}{2021}), \bibinfo{pages}{535–560}.
\newblock


\bibitem[\protect\citeauthoryear{Zhang, Wang, Zhao, He, Zhong, Yuan, and
  Wang}{Zhang et~al\mbox{.}}{2018}]%
        {allowableimp}
\bibfield{author}{\bibinfo{person}{S. Zhang}, \bibinfo{person}{W. Wang},
  \bibinfo{person}{H. Zhao}, \bibinfo{person}{F. He}, \bibinfo{person}{K.
  Zhong}, \bibinfo{person}{S. Yuan}, {and} \bibinfo{person}{Z. Wang}.}
  \bibinfo{year}{2018}\natexlab{}.
\newblock \showarticletitle{Status of internal quality control for thyroid
  hormones immunoassays from 2011 to 2016 in China}.
\newblock \bibinfo{journal}{\emph{Journel of Clinical Laboratory Analysis}}
  \bibinfo{volume}{32} (\bibinfo{date}{Jan.} \bibinfo{year}{2018}),
  \bibinfo{pages}{e22154}.
\newblock


\end{thebibliography}
